\documentclass[runningheads]{llncs}
\usepackage[T1]{fontenc}
\usepackage{graphicx,verbatim}
\usepackage{amsmath}
\usepackage{multirow}
\usepackage{amssymb}
\usepackage{booktabs}
\usepackage[colorlinks=true, urlcolor=blue, linkcolor=blue, citecolor=blue]{hyperref}
\usepackage[table]{xcolor}
\usepackage{marvosym} 
\usepackage{overpic}
\usepackage{rotating}
\usepackage{multirow}

\newcommand{\figref}[1]{Fig. \ref{#1}}

\definecolor{mygray}{gray}{.92}
\def\etal{{\em et al.}}

\def\etal{{\em et al.~}}

\newsavebox\CBox
\def\textBF#1{\sbox\CBox{#1}\resizebox{\wd\CBox}{\ht\CBox}{\textbf{#1}}}

\graphicspath{{./Imgs/}}
\DeclareGraphicsExtensions{.jpg,.pdf,.png}

\begin{document}

\title{Text-driven Multiplanar Visual Interaction for Semi-supervised Medical Image Segmentation}

\author{Kaiwen Huang\inst{1}, Yi Zhou\inst{2}, Huazhu Fu\inst{3}, Yizhe Zhang\inst{1}, Chen Gong\inst{1}, Tao Zhou\inst{1} \textsuperscript{(\Letter)}}


\authorrunning{Huang et al.}
\institute{
$^1$ School of Computer Science and Engineering, Nanjing University of Science and Technology, China.\\
{\tt taozhou.ai@gmail.com}\\
$^2$ School of Computer Science and Engineering, Southeast University, China. \\
$^3$ Institute of High Performance Computing, Agency for Science, Technology and Research, Singapore. \\
}

\titlerunning{Text-SemiSeg for semi-supervised segmentation}

%

\maketitle              
\begin{abstract}
Semi-supervised medical image segmentation plays a critical method in mitigating the high cost of data annotation. When labeled data is limited, textual information can provide additional context to enhance visual semantic understanding. However, research exploring the use of textual data to enhance visual semantic embeddings in 3D medical imaging tasks remains scarce. In this paper, we propose a novel text-driven multiplanar visual interaction framework for semi-supervised medical image segmentation (termed Text-SemiSeg), which consists of three main modules: Text-enhanced Multiplanar Representation (TMR), Category-aware Semantic Alignment (CSA), and Dynamic Cognitive Augmentation (DCA). Specifically, TMR facilitates text-visual interaction through planar mapping, thereby enhancing the category awareness of visual features. CSA performs cross-modal semantic alignment between the text features with introduced learnable variables and the intermediate layer of visual features. DCA reduces the distribution discrepancy between labeled and unlabeled data through their interaction, thus improving the model’s robustness. Finally, experiments on three public datasets demonstrate that our model effectively enhances visual features with textual information and outperforms other methods. Our code is available at \url{https://github.com/taozh2017/Text-SemiSeg}.

\end{abstract}

\section{Introduction}\label{sec:Introduction}

Semi-supervised medical image segmentation~\cite{gu2025dual,huang2025learnable,jiao2024learning}, which blends limited labeled data with extensive unlabeled data, has garnered increasing attention for alleviating manual annotation burdens. Existing semi-supervised methods are mainly divided into consistency learning~\cite{basak2022exceedingly,bortsova2019semi} and pseudo-labeling approaches~\cite{han2022effective,zhang2022discriminative}. Consistency learning enhances the model's generalization ability by ensuring consistent predictions across various perturbed inputs. For example, Huang~\etal\cite{huang2022semi} employed cutout and slice misalignment as input perturbations for consistency learning. Additionally, Xu~\etal\cite{xu2022learning} designed different network architectures to introduce perturbations at the feature level. Pseudo-labeling generates and iteratively refines pseudo-labels from model predictions on unlabeled data. Wang~\etal\cite{wang2022ssa} introduced a confidence block to assess the quality of pseudo-labels and used a thresholding to select high-confidence pseudo-labels. Han~\etal\cite{han2022effective} generated high-quality pseudo-labels by obtaining class prototypes from labeled data and calculating the feature distance similarity between unlabeled data and the class prototypes. However, these methods often exhibit a phenomenon where they achieve satisfactory results on a specific medical modality or organ but lack generalizability to other tasks.


Recent work based on Visual-Language Models~(VLMs), such as Contrastive Language-Image Pre-training (CLIP) \cite{radford2021learning}, has achieved remarkable results by aligning images with their corresponding text descriptions through contrastive learning. 
In the medical imaging domain, 
Liu~\etal\cite{liu2023clip} pioneered the application of CLIP to 3D medical imaging tasks, demonstrating that textual information can provide additional semantic context to supplement voxel-based visual features. Additionally, VCLIPSeg~\cite{li2024vclipseg} applies CLIP to semi-supervised 3D medical segmentation.
However, VCLIPSeg simply replicates the textual information to match the dimensionality of the 3D data, which does not align with the 2D-oriented training paradigm of CLIP. Moreover, CLIP lacks learnable parameters to adapt to this cross-dimensional operation from 2D to 3D. \textbf{Thus, how can a semi-supervised text-visual model be designed to better accommodate 3D medical image segmentation?}

To this end,  we propose a novel \textbf{Text-SemiSeg} framework, which integrates textual information into semi-supervised 3D medical image segmentation. Specifically, we propose a Text-enhanced Multiplanar Representation (TMR) strategy to fully leverage text-visual alignment capabilities from CLIP as it is pre-trained on a large number of 2D images, thereby enhancing the semantic representation of visual features. Additionally, we present a Category-aware Semantic Alignment (CSA) module, which aligns visual embeddings with text features to augment the model's category semantic awareness. To further mitigate the well-known distributional discrepancy between labeled and unlabeled data in semi-supervised tasks, we propose the Dynamic Cognitive Augmentation (DCA) module. This approach reduces these distributional disparities, resulting in improved model robustness. \textbf{Our contributions are highlighted as follows:}
\textbf{{\romannumeral 1})} We propose a novel framework that is more suitable for applying CLIP in 3D semi-supervised medical segmentation scenarios. More importantly, this framework can also be applied to other semi-supervised networks. 
\textbf{{\romannumeral 2})} We propose TMR and CSA to fully enhance visual features with textual cues. \textbf{{\romannumeral 3})} We present the DCA strategy to effectively reduce the distribution gap between labeled and unlabeled data.
\textbf{{\romannumeral 4})} Extensive experiments on three 3D medical datasets demonstrate that our model outperforms other state-of-the-art semi-supervised methods.


\begin{figure*}[!t]
	\centering
	\begin{overpic}[width=\textwidth]{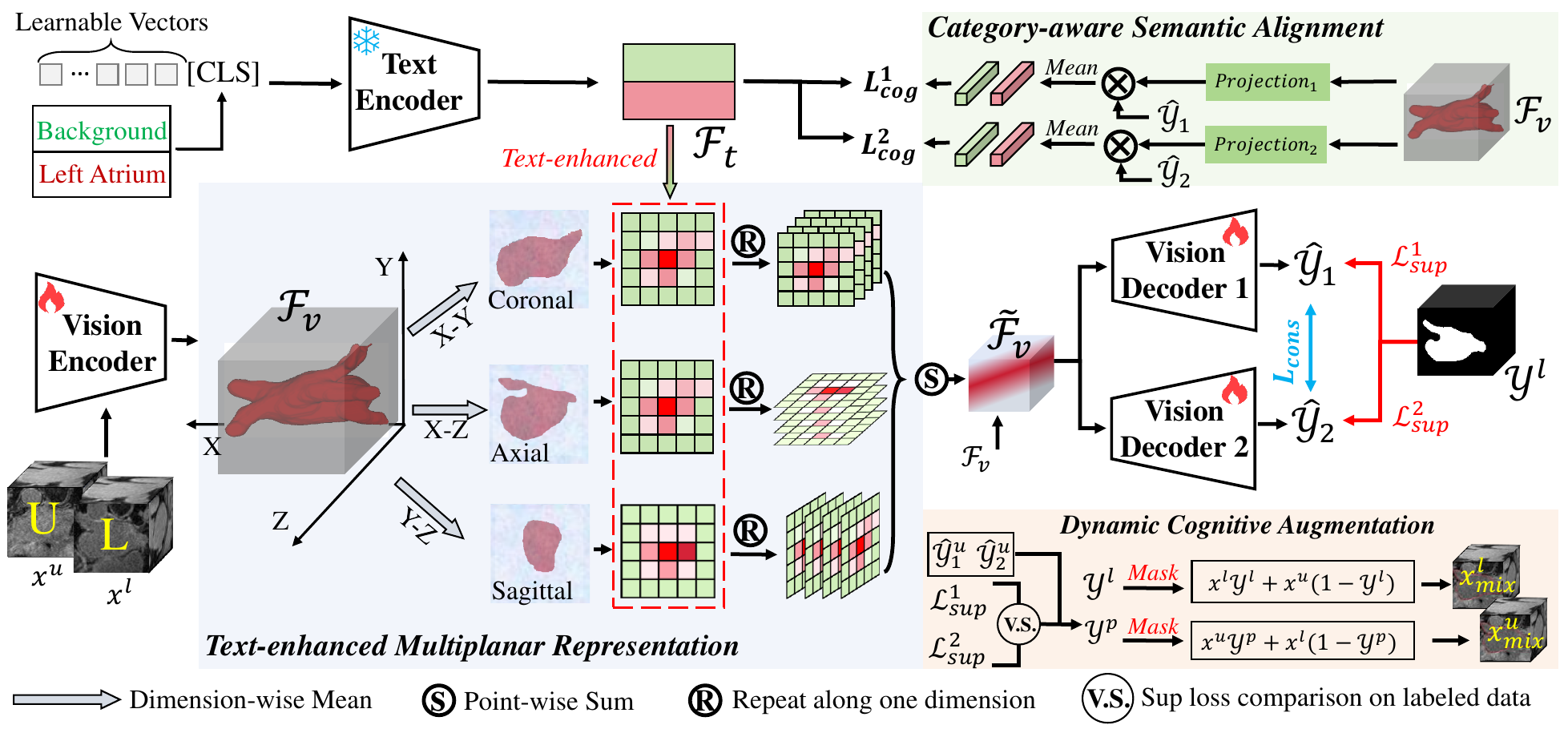}
    \end{overpic}
	\caption{{
    Overview of our Text-SemiSeg framework, including an MC-Net~\cite{wu2021semi} backbone and a text encoder. 
    The TMR and CSA leverage textual information to enhance visual feature learning, while the DCA adaptively generates mixed samples from both labeled and unlabeled data to reduce the distribution gap between them.
 }}
    \label{fig:Model}
\end{figure*}

\section{Proposed Method}\label{sec:Methods}

\textbf{Overview}. 
In the semi-supervised setting, the dataset comprises a labeled subset $\mathcal{D}^L=\{(x^l_i,y^l_i)\}^{N_l}_{i=1}$ and an unlabeled subset $\mathcal{D}^U=\{x^u_i\}^{N_l+N_u}_{i=N_l+1}$, where $N_l\ll{N_u}$. $x_i \in \mathbb{R}^{H \times W \times D}$ represents the input volume, and $y_i \in \{0,1\}^{H\times W \times D}$ is the corresponding ground-truth map. 
\figref{fig:Model} illustrates the architecture of our Text-SemiSeg framework. 
Specifically, the input volume $\{x_i^l, x_i^u\}$ is fed into the encoder to obtain feature embeddings. Meanwhile, the category text is incorporated in a learnable manner~\cite{zhou2022learning}, yielding text features $\mathcal F_t \in \mathbb{R}^{K \times C}$, where $K$ and $C$ represent the number of categories and channels, respectively. 
In the TMR module, to align with the CLIP training paradigm, we extract feature embeddings from distinct planes, generating three 2D features. We then enhance the visual features of each perspective by incorporating textual information. Subsequently, the enhanced feature embeddings are fed into two decoders for consistency learning. Then, we leverage the CSA module to facilitate the alignment of textual and visual features in their representations. Finally, we incorporate the DCA to reduce the distributional discrepancies between labeled and unlabeled data. 

\subsection{Text-enhanced Multiplanar Representation}\label{sec:Text-Enhanced Multiplanar Embedding}

CLIP is a text-visual model trained through contrastive learning between 2D images and text. However, the transformation of pixel-level semantics into voxel-level semantic information is still challenging. 
To address this issue, we employ two strategies: (1) incorporating learnable variables into the text to enhance the generalizability of text features, and (2) performing semantic enhancement between text features and various planes of 3D feature embeddings. 
Specifically, the text description in CLIP no longer uses the format \texttt{`A photo of [CLS]'}. Instead, we employ a prompt learning approach~\cite{zhou2022learning} as follows:
\begin{equation} 
\mathcal {F}_t = [V_1][V_2]\dots[V_M][CLS],
\end{equation}
where $[V]$ and $[CLS]$ represent learnable variables and class embeddings, respectively. The input volume is fed into the vision encoder to obtain visual embeddings $\mathcal F_v$.
Subsequently, the visual embeddings are compressed into 2D images via adaptive average pooling along the width, height, and depth dimensions, respectively. This process yields feature embeddings for the coronal plane $\mathcal F_v^c$, sagittal plane $\mathcal F_v^s$, and axial plane $\mathcal F_v^a$. These planes then interact with the text features through a text-enhanced attention operation. 
For the coronal plane, self-attention is initially applied. Then, text enhancement is performed on the plane, which can be expressed as:
\begin{equation} 
\tilde {\mathcal {F}_v^c} = \sigma\left(\frac{\mathbf{Q}(\mathcal {F}_t) \cdot \mathbf{K}({\mathcal A( \mathcal F}_v^c))^\top}{\sqrt{d}}\right) \cdot \mathbf{V}({\mathcal A (\mathcal F}^c_v)),
\end{equation}
where $\mathcal {A}(\cdot)$ denotes the attention operation.
$\mathbf{Q}$, $\mathbf{K}$, and $\mathbf{V}$ represent the linear projections. $\sqrt{d}$ is a scaling factor, and $\sigma(\cdot)$ represents the softmax activation function. 
For the sagittal and axial planes, the operations are consistent with those detailed above. Finally, the voxel embeddings are reconstructed from the three text-enhanced planes, and the process can be mathematically expressed as
\begin{equation}
\tilde {\mathcal F_v}(x, y, z) = \sum_{x=1}^H \sum_{y=1}^W \sum_{z=1}^D \left( w_c \cdot \tilde {\mathcal F^c_v}(x, y) + w_s \cdot \tilde {\mathcal F^s_v}(y, z) + w_a \cdot \tilde {\mathcal F^a_v}(x, z)\right) + \mathcal F_v,
\end{equation}
where $w_c$, $w_s$, and $w_a$ are learnable weights that dynamically regulate the contributions of three different planes in the reconstruction of 3D voxel features. 


\subsection{Category-aware Semantic Alignment}
\label{sec:Vision-Text Cognitive Alignment}


We leverage the proposed CSA module to enhance the network's contextual understanding of text-image pairs. 
Specifically, the predictions of the two visual decoders are multiplied by the final layer features of the visual encoder to obtain the corresponding category semantic representations. We apply MSE regularization to constrain each category’s text feature to be as close as possible to the visual embedding, which can be expressed as:
\begin{equation}
\mathcal L_{cog} = \sum\nolimits_{k=1}^K (\mathcal F_t^k -  \text {avg}( \mathcal F_v^p \cdot \hat{y}_{1,k}))^2 + (\mathcal F_t^k -  \text {avg}(\mathcal F_v^p \cdot \hat{y}_{2,k}))^2 ,
\end{equation}
where $\mathcal F_t^k $ represents the text feature of the  $k$-th category. $\hat{y}_{1,k}$ and $\hat{y}_{2,k}$ represent the predictions of the two visual decoders for the $k$-th category, respectively. $\text {avg}(\cdot)$ denotes the global average pooling operation. $\mathcal F_v^p = {proj}(\mathcal F_v)$ represents the output of the visual feature $\mathcal F_v$ passing through the projection head, which ensures that it matches the dimensionality of the text features.

\subsection{Dynamic Cognitive Augmentation}
\label{sec:Dynamic Cognitive Retraining Augmentation}

In semi-supervised tasks, due to the scarcity of labeled data, there is often a distributional discrepancy between labeled and unlabeled data. Reducing this discrepancy can enhance the model's ability to recognize the foreground and reduce the impact of various types of noise. 
Thus, we use the GT and pseudo-labels to extract the foreground parts from both labeled and unlabeled data and fuse them into each other's background. This process can be expressed as:
\begin{equation}
\left\{
\begin{aligned}
& x_{mix}^l = \sum_{k=1}^K x^l \cdot y^l_k + x^u \cdot  (1 - y^l_k), 
& x_{mix}^u = \sum_{k=1}^K x^u \cdot y^p_k + x^l \cdot  (1 - y^p_k), \\
& y^p = \begin{cases} 
      \mathcal B(\hat{y}^u_1),  &  \mathcal L_{sup}^1 < \mathcal L_{sup}^2, \\
      \mathcal B(\hat{y}^u_2), & \mathcal L_{sup}^1 > \mathcal L_{sup}^2,
   \end{cases}
\end{aligned}
\right.
\end{equation}
where $\mathcal B(\cdot)$ represents the binarization operation. $y^l_k$ and $y^p_k$ denote the $k$-th class ground truth and pseudo-label, respectively. $\hat{y}^u_1$ and $\hat{y}^u_2$ represent the predictions of the two decoders for the unlabeled data, respectively.
In the process of generating pseudo-labels, we select the decoder output with the lower loss on the labeled data as the pseudo-label for the unlabeled data. Then, when the mixed data is fed into the model for retraining, loss constraints are applied only to the underperforming decoder, encouraging its improvement. The process for obtaining mixed labels can be expressed as follows:
\begin{equation}
\begin{aligned}
& y_{mix}^l = \sum\nolimits_{k=1}^K y^l \cdot y^l_k + y^p \cdot  (1 - y^l_k),  &y_{mix}^u = \sum\nolimits_{k=1}^K y^p \cdot y^p_k + y^l \cdot  (1 - y^p_k).\\
\end{aligned}
\end{equation}


\subsection{Overall Loss Function}\label{sec:Overall Loss Function}

In thi study, the segmentation loss consists of the Dice and cross-entropy (CE) losses, which can be defined by $\mathcal L_{seg}(\hat{y}, y ) = \mathcal L_{dice}(\hat{y}, y )  + \mathcal L_{ce}(\hat{y}, y )$, where $\hat{y}$ and $y$ denote the prediction and ground truth, respectively. Thus, we have
\begin{equation}
\begin{aligned}
& \mathcal L_{sup} = \mathcal L_{seg}(\hat{y}_1^l, y^l) + \mathcal L_{seg}(\hat{y}_2^l, y^l), 
& \mathcal L_{unsup} = \mathcal L_{seg}(\hat{y}_1^u, \hat{y}_2^u) + \mathcal L_{seg}(\hat{y}_2^u, \hat{y}_1^u).\\
\end{aligned}
\end{equation}

Moreover, the mixed loss $\mathcal L_{mix}$, which updates only the decoder with poorer performance in each iteration, can be expressed as follows:
\begin{equation}
\mathcal L_{mix} =  \mathcal L_{seg}(\hat{y}_{mix}^l, y_{mix}^l) + \mathcal L_{seg}(\hat{y}_{mix}^u, y_{mix}^u),
\end{equation}
where $\hat{y}_{mix}^l$ and $\hat{y}_{mix}^u$ represent the model's prediction on ${x}_{mix}^l$ and ${x}_{mix}^u$.

Finally, the overall loss function can be formulated by
\begin{equation}
\mathcal L_{total} = \mathcal L_{sup}  + \mathcal L_{cog} + \mathcal L_{mix} + \lambda_u \mathcal L_{unsup},
\end{equation}
where $\lambda_u$ represents Gaussian warm-up function~\cite{laine2016temporal}. 

\begin{figure*}[!t]
	\centering
	\begin{overpic}[width=\textwidth]{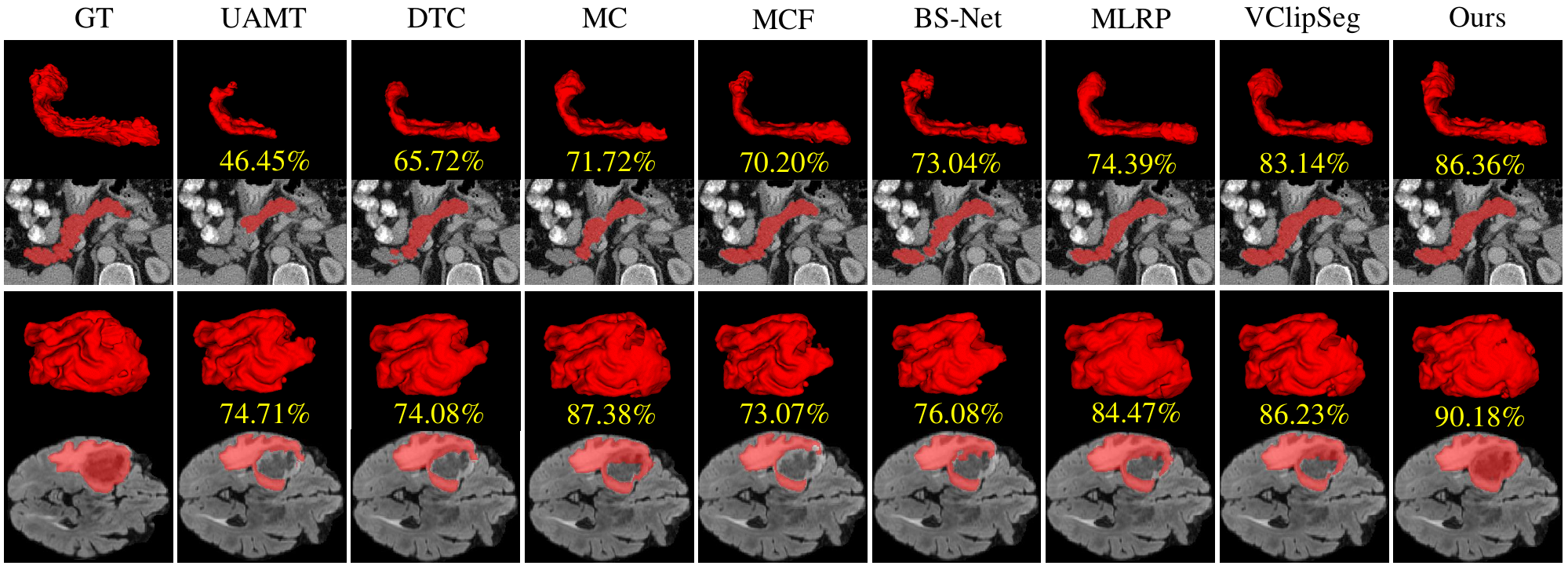}
    \end{overpic}
	\caption{{Visualization results on the Pancreas and BraTS-2019 datasets, with yellow numbers indicating the dice value of the sample.  
 }}
    \label{fig:res}
\end{figure*}

\section{Experiments and Results}\label{sec:Experiments}

\subsection{Experimental Setup}

\textbf{Datasets}.
We conduct comparative experiments on the
Pancreas-CT~\cite{clark2013cancer}, Brats-2019~\cite{hdtd-5j88-20} and MSD-Lung\cite{antonelli2022medical} tumor datasets.
$\bullet$ The Pancreas-CT dataset includes 82 3D abdominal CT scans with a resolution of $512\times512$ and slice thickness of $1.5$-$2.5$ $mm$. We use 62 samples for training and 20 for testing, based on the same setting in~\cite{luo2021semi}. 
$\bullet$ The BraTS-2019 dataset consists of MRI images from 335 glioma patients, with four modalities: T1, T2, T1CE, and FLAIR. Following~\cite{ssl4mis2020} setup, we use only the FLAIR images for tumor segmentation. The data is split into 250 samples for training, 25 for validation, and 60 for testing.
$\bullet$ The MSD-Lung dataset consists of 63 CT scans. Following Setting \cite{zeng2025pick}, we allocated 48 scans for training and 15 for testing.

\noindent\textbf{Implementation Details.} 
All experiments are conducted in a PyTorch 1.8.1 environment with CUDA 11.2 and an NVIDIA 4090 GPU. 
To ensure fairness, the backbone selection continues to follow previous settings, choosing VNet~\cite{milletari2016v}. 
We utilize the SGD optimizer with an initial learning rate of $0.01$. The batch size is set to $4$, and the number of iterations is set to $30$k. 
The number of learnable parameters in CLIP is set to 4.
We use a Gaussian warm-up function $\lambda_{m}(t) = \beta * e^{-5(1-t/t_{max})^2}$ in $\mathcal L_{unsup}$, where $\beta$ is set to $0.1$.
Following the previous settings~\cite{su2024mutual}, during training, we randomly extract 3D patches of size $96 \times 96 \times 96$.

\begin{table*}[t!]
  \centering
  \scriptsize
  \renewcommand{\arraystretch}{1.15}
  \setlength\tabcolsep{0.8pt}
  \caption{Quantitative results on the Pancreas, BraTS-2019 and MSD Lung datasets.
  }\label{tab:tab_Comparison}
\begin{tabular}{c||rr|cccc|cccc}

\hline
{\multirow{2}{*}{Dataset}} &\multicolumn{2}{c|}{\multirow{2}{*}{Method}}  & \multicolumn{4}{c|}{10\% / 6 labeled data} & \multicolumn{4}{c}{20\% / 12 labeled data} \\
\cline{4-11}
   & & & Dice~$\uparrow$ & Jaccard~$\uparrow$ & 95HD~$\downarrow$ & ASD~$\downarrow$    & Dice~$\uparrow$ & Jaccard~$\uparrow$ & 95HD~$\downarrow$ & ASD~$\downarrow$    \\
\hline                      
\multirow{12}{*}{\begin{sideways}Pancreas\end{sideways}} &\multicolumn{2}{c|}{VNet~\cite{milletari2016v}~(SupOnly)}     &   54.94& 40.87  &  47.48 & 17.43 &   75.07  &   61.96& 10.79& 3.31\\
\cline{2-11}

&UAMT~\cite{yu2019uncertainty}    & {\fontsize{1pt}{5pt}\selectfont\textcolor{black}{(MICCAI'19)}} &  66.44 &  52.02   &  17.04       &   3.03  &      76.10   & 62.62      &    10.87    &   2.43      \\
&DTC~\cite{luo2021semi} & {\fontsize{1pt}{5pt}\selectfont\textcolor{black}{(AAAI'21)}}  & 66.58   &  51.79   &   15.46  & 4.16  &  76.27   &   62.82  & 8.70 & 2.20 \\
&MC-Net~\cite{wu2021semi}    &{\fontsize{1pt}{5pt}\selectfont\textcolor{black}{(MICCAI'21)}} & 69.07 & 54.36 &    14.53     &   2.28     &   78.17         &     65.22 &    6.90     &  1.55    \\
&MC-Net+ \cite{wu2022mutual} &{\fontsize{1pt}{5pt}\selectfont\textcolor{black}{(MIA'22)}} & 70.00 &  55.66  & 16.03  & 3.87 & 79.37  &  66.83    & 8.52 & 1.72 \\
&URPC \cite{luo2022semi} &{\fontsize{1pt}{5pt}\selectfont\textcolor{black}{(MIA'22)}} &  73.53 &   59.44 &  22.57  & 7.85 &    80.02   & 67.30  &   8.51  &  1.98  \\
&MCF~\cite{wang2023mcf}   &{\fontsize{1pt}{5pt}\selectfont\textcolor{black}{(CVPR'23)}}  & 67.71 &  53.83 &  17.17  &  {2.34} &  75.00 &  61.27 &  11.59  &  3.27 \\
&BCP~\cite{bai2023bidirectional}   &{\fontsize{1pt}{5pt}\selectfont\textcolor{black}{(CVPR'23)}}  & 75.17   &  60.99 &  10.49 & 2.32       &  {82.91} &  {70.97} & {6.43} & 2.25 \\
&BS-Net~\cite{he2023bilateral}  &{\fontsize{1pt}{5pt}\selectfont\textcolor{black}{(TMI'24)}}  &  64.61        &  50.02  &  24.74  &  5.28 & 78.93   & 65.75       &8.49  &  2.20  \\

&MLRP~\cite{su2024mutual}  &{\fontsize{1pt}{5pt}\selectfont\textcolor{black}{(MIA'24)}}  & {75.93} &  {62.12}   &  {9.07}   &  {1.54} & {81.53} & 69.35  &  6.81 & {1.33}\\
&VClipSeg~\cite{li2024vclipseg} &{\fontsize{1pt}{5pt}\selectfont\textcolor{black}{(MICCAI'24)}}  &  74.77  & 60.88     & 11.02 & {1.82} &  80.56 & 68.13 & 6.75 &  1.49\\

&\textbf{Ours} & &    \textBF{81.27}      &   \textBF{68.78}   & \textBF{5.98}   &  \textBF{1.44}  &    \textBF{83.50}  &  \textBF{71.94}   &   \textBF{4.52} &  \textBF{1.26}      \\

\hline

\multirow{13}{*}{\begin{sideways}BraTS-2019\end{sideways}} &\multicolumn{2}{c|}{Method} & \multicolumn{4}{c|}{10\% / 25 labeled data} & \multicolumn{4}{c}{20\% / 50 labeled data}\\
\cline{2-11}
&\multicolumn{2}{c|}{VNet~\cite{milletari2016v}~(SupOnly)}     &  74.43  &  61.86  &   37.11   &  2.79      &  80.16        &    71.55 &   22.68 &   3.43 \\
\cline{2-11}
&UAMT~\cite{yu2019uncertainty}    & {\fontsize{1pt}{5pt}\selectfont\textcolor{black}{(MICCAI'19)}} &  79.49  &   69.22  &  11.93  &  1.93 & 79.72  & 69.46 & 11.26  & 1.97        \\
&DTC~\cite{luo2021semi} & {\fontsize{1pt}{5pt}\selectfont\textcolor{black}{(AAAI'21)}}  & 77.54   &  66.91   & 12.16  & 2.84  &  82.38  &  72.15  &  11.00 & 2.16 \\
&MC-Net~\cite{wu2021semi}    &{\fontsize{1pt}{5pt}\selectfont\textcolor{black}{(MICCAI'21)}} & 81.52   &  71.89  &  13.26  &  4.56 & 83.79  &   73.69  &  9.65 & 1.76   \\
&MC-Net+ \cite{wu2022mutual} &{\fontsize{1pt}{5pt}\selectfont\textcolor{black}{(MIA'22)}}  &   79.27  &   68.97& 14.89  & 4.91  & 83.47 &  72.89  &  9.69  & 1.92 \\
&URPC \cite{luo2022semi} &{\fontsize{1pt}{5pt}\selectfont\textcolor{black}{(MIA'22)}} &   82.59  &  72.11     &  13.88    &  3.72       &  82.93   & 72.57  & 15.93   &  4.19  \\
&MCF~\cite{wang2023mcf}   &{\fontsize{1pt}{5pt}\selectfont\textcolor{black}{(CVPR'23)}} &     78.83     &   68.49   &   12.25    &  1.92  &  80.07 &  69.55 &  10.65  & 2.00  \\
&BCP~\cite{bai2023bidirectional}   &{\fontsize{1pt}{5pt}\selectfont\textcolor{black}{(CVPR'23)}}& 78.64    & 68.59   &  12.88  &2.81  & 80.20  &  69.66 &  10.52 &  2.08     \\
&BS-Net~\cite{he2023bilateral}  &{\fontsize{1pt}{5pt}\selectfont\textcolor{black}{(TMI'24)}}  &  79.93 &  69.37  & 10.88  & 2.05 &    82.03  &   71.87   &  10.26 &  1.96\\
&MLRP~\cite{su2024mutual}  &{\fontsize{1pt}{5pt}\selectfont\textcolor{black}{(MIA'24)}}  &  {84.29}  & {74.74}  &  {9.57}  & 2.55 &  {85.47}      & {76.32} &  {7.76} & 2.00 \\
&VClipSeg~\cite{li2024vclipseg} &{\fontsize{1pt}{5pt}\selectfont\textcolor{black}{(MICCAI'24)}}& 83.12     & 73.61   & 10.40  &  {1.65}  &  84.05 & 74.38   & 9.75 & {1.35}   \\
&\textbf{Ours} & & \textBF{85.27}  & \textBF{75.83}  &  \textBF{8.77} & \textBF{1.35}  & \textBF{86.69}  & \textBF{77.83}     & \textBF{7.65}   &  \textBF{1.24}  \\

\hline

\multirow{13}{*}{\begin{sideways}MSD-Lung\end{sideways}} &\multicolumn{2}{c|}{Method} & \multicolumn{4}{c|}{20\% / 10 labeled data} & \multicolumn{4}{c}{40\% / 20 labeled data}\\
\cline{2-11}
&\multicolumn{2}{c|}{VNet~\cite{milletari2016v}~(SupOnly)}     &  36.36 &   25.78 & 19.06 & 32.73  &  57.67  &  43.24   &  11.99   &   4.62  \\
\cline{2-11}
&UAMT~\cite{yu2019uncertainty}    & {\fontsize{1pt}{5pt}\selectfont\textcolor{black}{(MICCAI'19)}} &   40.22       & 28.10      &  16.79 & 7.02  &  61.24 &   46.51   & 8.97  & 3.54 \\

&DTC~\cite{luo2021semi} & {\fontsize{1pt}{5pt}\selectfont\textcolor{black}{(AAAI'21)}}  &  59.64 & 44.23   &  9.00  & 2.89   &  63.40   & 48.47      &   12.85  &  7.82 \\

&MC-Net~\cite{wu2021semi}    &{\fontsize{1pt}{5pt}\selectfont\textcolor{black}{(MICCAI'21)}} & 52.23 &  39.09  &  13.85  & 7.05   &  56.10  &  43.69   &   16.87    &  10.92   \\

&MC-Net+ \cite{wu2022mutual} &{\fontsize{1pt}{5pt}\selectfont\textcolor{black}{(MIA'22)}} &  58.72  &  44.47   & 10.51  &  4.15  &  59.52  &   47.25 & 12.12  &  5.87 \\

&URPC \cite{luo2022semi} &{\fontsize{1pt}{5pt}\selectfont\textcolor{black}{(MIA'22)}} &   55.19  &  39.80    & 9.08 &  {2.42} &  62.20  &   47.93   &  {8.22} & \textBF{2.38}  \\
&MCF~\cite{wang2023mcf}   &{\fontsize{1pt}{5pt}\selectfont\textcolor{black}{(CVPR'23)}}  &   53.81 &  39.23 &  12.41 & 4.35  &  62.86  &  48.62    &   8.42 &  3.67 \\

&BCP~\cite{bai2023bidirectional}   &{\fontsize{1pt}{5pt}\selectfont\textcolor{black}{(CVPR'23)}}  & 59.34  &  44.83  &  11.12 & 3.69  & 65.01 &  48.70  & 9.79 & 3.47 \\
&BS-Net~\cite{he2023bilateral}  &{\fontsize{1pt}{5pt}\selectfont\textcolor{black}{(TMI'24)}}  &  47.33  &  32.78  &  9.54 & 2.92  &  53.30 & 40.02  & 11.71  & 4.33 \\
&MLRP~\cite{su2024mutual}  &{\fontsize{1pt}{5pt}\selectfont\textcolor{black}{(MIA'24)}}  &  {62.12}  & {47.23}  & {8.47}  &  2.68  & {67.63}  &   {53.35}    &   12.30 &  7.56 \\
&VClipSeg~\cite{li2024vclipseg} &{\fontsize{1pt}{5pt}\selectfont\textcolor{black}{(MICCAI'24)}}  &  59.39  &  44.55  &  8.81  &  2.68  & 67.58  &   53.19 &  12.38 &  7.64 \\
&\textbf{Ours} & &  \textBF{65.02}  &  \textBF{48.90}  & \textBF{6.25} &  \textBF{1.28}  & \textBF{68.21}   & \textBF{54.02} & \textBF{8.08} & {2.97} \\

\hline
\end{tabular}
\end{table*}

\subsection{Comparison with State-of-the-arts}

We compare the proposed model with ten state-of-the-art semi-supervised medical image segmentation methods, namely 
UA-MT~\cite{yu2019uncertainty}, DTC~\cite{luo2021semi}, MC-Net~\cite{wu2021semi}, MC-Net+~\cite{wu2022mutual}, URPC~\cite{luo2022semi}, MCF~\cite{wang2023mcf}, BCP~\cite{bai2023bidirectional}, BS-Net~\cite{he2023bilateral}, MLRP~\cite{su2024mutual}, and VClipSeg~\cite{li2024vclipseg}. Table~\ref{tab:tab_Comparison} presents a quantitative comparison and our method achieves the best performance across three datasets. Specifically, under the condition of $10\%$ labeled data, our model surpasses the second-best method by $5.34\%$, $0.98\%$, and $2.90\%$ in Dice coefficient on the Pancreas, BraTS, and Lung datasets, respectively. 
\figref{fig:res} illustrates the qualitative comparison with $10\%$ labeled data. 
It can be observed that the proposed framework, Text-SemiSeg, effectively enhances visual feature representation through textual information, leading to more precise predictions, particularly in the edge regions.


\subsection{Ablation Studies}

$\bullet$ \noindent \textbf{Effectiveness of Key Components:} 
We used the MC-Net~\cite{wu2021semi} structure as a baseline and conducted ablation experiments on the proposed TMR, CSA, and DCA modules, as shown in Table \ref{tab:tab_Abi_Block}. From the results, it can be observed that each component contributes the promising performance. 

\begin{table*}[t!]
  \centering
  \scriptsize
  \renewcommand{\arraystretch}{1.15}
  \setlength\tabcolsep{2.8pt}
  \caption{Ablation study on the key components using 20\% labeled data.
  }\label{tab:tab_Abi_Block}
\begin{tabular}{ccc|cccc|cccc}
\hline

\multicolumn{3}{c|}{Settings} & \multicolumn{4}{c|}{Pancreas~(12 labeled data)} & \multicolumn{4}{c}{BraTS-2019}~(50 labeled data) \\
\hline
TMR     & CSA     & DCA     & Dice~$\uparrow$   & IoU~$\uparrow$  & 95HD~$\downarrow$  & ASD~$\downarrow$  & Dice~$\uparrow$   & IoU~$\uparrow$   & 95HD~$\downarrow$   & ASD~$\downarrow$  \\
\hline
&    &         &   78.17   & 65.22  & 6.90    &  1.55    &   83.79 &  73.69 &  9.65  & 1.76  \\
\checkmark     &    &         & 79.46   & 67.11   & 8.30 & 1.20      &  84.63  &  75.02     &   9.41     &  1.60    \\

&    \checkmark      &   &    80.26    &   67.60   &   8.96    &   1.26   &     84.72   &   75.31    &    9.39    &   1.63   \\

&         &    \checkmark      &  83.41  & 71.81     & 4.44      & 1.19     &    85.38    & 76.17  &   9.05     & 1.26  \\

\checkmark    &    \checkmark     &    &  82.10 &  69.96 &  5.32  &  \textBF{1.14}     &  85.19 &  75.79 & 9.32    &  1.54    \\

\checkmark   &         &   \checkmark       &  83.47  & 71.85   &   4.59    &{1.20}   &  86.30  & 77.11 &  9.10  &  1.58\\

&    \checkmark      &  \checkmark        &   83.28   & 71.63   &  4.59   &  1.29    &  85.89   &  76.56 &  8.77  &  1.65  \\
\checkmark &    \checkmark  &  \checkmark   &   \textBF{83.50}  & \textBF{71.94}   &  \textBF{4.52} & 1.26      &  \textBF{86.69}  & \textBF{77.83}   & \textBF{7.65}  & \textBF{1.24}   \\

\hline

\end{tabular}
\end{table*}

\begin{table*}[t!]
  \centering
  \scriptsize
  \renewcommand{\arraystretch}{1.2}
  \setlength\tabcolsep{3.8pt}
  \caption{Ablation study of the text-vision enhancement strategy.
  }\label{tab:tab_Abi_Multiplanar}
\begin{tabular}{c|cccc|cccc}
\hline
\multirow{2}{*}{Settings} & \multicolumn{4}{c|}{Pancreas~(12 labeled data)} & \multicolumn{4}{c}{BraTS-2019}~(50 labeled data) \\
\cline{2-9}
& Dice~$\uparrow$   & IoU~$\uparrow$  & 95HD~$\downarrow$  & ASD~$\downarrow$  & Dice~$\uparrow$   & IoU~$\uparrow$   & 95HD~$\downarrow$   & ASD~$\downarrow$  \\
\hline
Repeat &  82.35      &    70.42  &   7.15    &  2.03    &  85.57   &   76.16    &    8.50    &   1.33   \\

Multiplanar &   \textBF{83.50}  & \textBF{71.94}   &  \textBF{4.52} & \textBF{1.26}      &  \textBF{86.69}  & \textBF{77.83}  &  \textBF{7.65} & \textBF{1.24}     \\
\hline
\end{tabular}
\end{table*}

$\bullet$ \noindent \textbf{Effects of Multiplanar Strategy:}
To demonstrate the effectiveness of adopting a multiplanar strategy, we compare it to the approach used by VClipSeg, which replicates textual features to match the dimensionality of the visual features. As shown in Table \ref{tab:tab_Abi_Multiplanar}, the Dice coefficients improve from $82.35\%$ and $85.57\%$ to $83.50\%$ and $86.69\%$ on the Pancreas and BraTS datasets, respectively. The multiplanar strategy offers several advantages: (1) it is more aligned with CLIP’s training paradigm, and (2) it can achieve global text-vision enhancement for 2D planar features using only a small number of parameters.

\begin{table*}[t!]
  \centering
  \scriptsize
  \renewcommand{\arraystretch}{1.2}
  \setlength\tabcolsep{5.0pt}
  \caption{Effect of text-enhanced strategy applied to other semi-supervised methods.
  }\label{tab:tab_Abi_Fairness}
\begin{tabular}{l|cccc|cccc}
\hline
\multirow{2}{*}{Settings} & \multicolumn{4}{c|}{Pancreas~(12 labeled data)} & \multicolumn{4}{c}{BraTS-2019}~(50 labeled data) \\
\cline{2-9}
& Dice~$\uparrow$   & IoU~$\uparrow$  & 95HD~$\downarrow$  & ASD~$\downarrow$  & Dice~$\uparrow$   & IoU~$\uparrow$   & 95HD~$\downarrow$   & ASD~$\downarrow$  \\
\hline
UAMT~\cite{yu2019uncertainty} &   76.10     &  62.62    &    10.87   &  2.43    &  79.72   &  69.46     &   11.26     &  1.97    \\

+Text&    \textBF{77.53}    &  \textBF{64.57}    &   \textBF{10.54}    &   \textBF{1.16}   &  \textBF{84.21}   & \textBF{74.86}  & \textBF{9.49}   & \textBF{1.65}     \\
\hline
DTC~\cite{luo2021semi}&   76.27    &  62.82    &   8.70    &  2.20  & 82.38  &  72.15   & 11.00   & 2.16     \\
+Text& \textBF{81.04}  & \textBF{68.70}   & \textBF{7.74}    &  \textBF{1.27}    &   \textBF{84.97} & \textBF{75.48}   &  \textBF{9.14} &   \textBF{1.66}   \\

\hline
\end{tabular}
\end{table*}

$\bullet$ \noindent \textbf{Fairness Discussion:} Currently, the exploration of VLMs in 3D scenarios remains quite limited. While some comparative methods in the experiments do not rely on text, the inference cost of Text-SemiSeg is comparable to other semi-supervised methods. The primary focus of this work is to explore how to better apply VLMs to downstream tasks. Table \ref{tab:tab_Abi_Fairness} presents the results of incorporating the text-enhanced strategy into other semi-supervised methods, while excluding other strategies such as DCA. The results clearly demonstrate that our text-enhanced strategy effectively boosts segmentation accuracy across different methods.




\section{Conclusion}\label{conclusion}

We have proposed a novel Text-SemiSeg framework for semi-supervised 3D medical image segmentation. Our model fully leverages CLIP's alignment capabilities to enrich visual features with textual cues in a 3D context. Additionally, to enhance the interaction between labeled and unlabeled data, we propose dynamic cognitive augmentation, which significantly boosts the model's generalization ability. Extensive experiments across various segmentation tasks show that our model outperforms existing methods and adapts more effectively to 3D scenarios.

\bibliographystyle{splncs03}
\bibliography{refs}

\end{document}